\documentclass[runningheads]{llncs}

\usepackage[numbers,square,sort&compress]{natbib}
\usepackage[skip=0pt]{caption}
\usepackage{graphicx}
\usepackage{color}
\usepackage{xcolor}
\usepackage{mathtools}
\usepackage{multirow}
\usepackage{colortbl}
\usepackage{comment}
\usepackage{accents}
\usepackage{hyperref}
\hypersetup{
  colorlinks   = true,    
  urlcolor     = blue,    
  linkcolor    = blue,    
  citecolor    = blue      
}
\usepackage{enumerate}
\usepackage[shortlabels,inline]{enumitem}
\usepackage{booktabs}
\usepackage{tikz}
\usepackage{hyperref}
\usepackage{amsmath}
\usepackage{url}
\usepackage{dsfont}
\usepackage[para]{footmisc}

\DeclarePairedDelimiter\abs{\lvert}{\rvert}

\newcommand*\circled[1]{\tikz[baseline=(char.base)]{
    \node[shape=circle,fill,inner sep=2.5pt] (char) {\textcolor{white}{#1}};}}
\definecolor{ballblue}{rgb}{0.13, 0.67, 0.8}
\definecolor{caribbeangreen}{rgb}{0.0, 0.8, 0.6}
\definecolor{cherryblossompink}{rgb}{1.0, 0.72, 0.77}
\definecolor{fluorescentorange}{rgb}{1.0, 0.75, 0.0}
\definecolor{skyblue}{rgb}{0.53, 0.81, 0.92}
\definecolor{stildegrainyellow}{rgb}{0.98, 0.85, 0.37}
\definecolor{silver}{rgb}{0.753,0.753,0.753}
\newcommand{\header}[1]{\vspace*{1mm}\noindent\textbf{#1}.}
\newcommand{\wip}[1]{\textbf{[WIP]}}

\author{
    Amin Abolghasemi\inst{1}
    \and
    Leif Azzopardi\inst{2}
    \and
    Arian Askari\inst{1}
    \and \\
    Maarten de Rijke\inst{3}
    \and
    Suzan Verberne\inst{1}
    \institute{
    Leiden University, The Netherlands
    \and
    University of Strathclyde, UK
    \and
    University of Amsterdam, The Netherlands
    \\
    \email{
    m.a.abolghasemi@liacs.leidenuniv.nl,
    leif.azzopardi@strath.ac.uk,
    a.askari@liacs.leidenuniv.nl,
    m.derijke@uva.nl,
    s.verberne@liacs.leidenuniv.nl,
    }
    }
}

\authorrunning{A. Abolghasemi et al.}

\begin{document}
\title{Measuring Bias in a Ranked List \\using Term-based Representations}

\maketitle              

\vspace{-4mm}
\begin{abstract}
In most recent studies, gender bias in document ranking is evaluated with the NFaiRR metric, which measures bias in a ranked list based on an aggregation over the unbiasedness scores of each ranked document. This perspective in measuring the bias of a ranked list has a key limitation: individual documents of a ranked list might be biased while the ranked list as a whole balances the groups’ representations. 
To address this issue, we propose a novel metric called TExFAIR (term exposure-based fairness), which is based on two new extensions to a generic fairness evaluation framework, attention-weighted ranking fairness (AWRF). TExFAIR assesses fairness based on the term-based representation of groups in a ranked list: (i) an explicit definition of associating documents to groups based on probabilistic term-level associations, and (ii) a rank-biased discounting factor (RBDF) for counting non-representative documents towards the measurement of the fairness of a ranked list.
We assess TExFAIR on the task of measuring gender bias in passage ranking, and study the relationship between TExFAIR and NFaiRR. Our experiments show that there is no strong correlation between TExFAIR and NFaiRR, which indicates that TExFAIR measures a different dimension of fairness than NFaiRR. With TExFAIR, we extend the AWRF framework to allow for the evaluation of fairness in settings with term-based representations of groups in documents in a ranked list.
\keywords{Bias \and Evaluation \and Document Ranking}
\end{abstract}

\section{Introduction}
\label{sec:introduction}

Ranked result lists generated by ranking models may incorporate biased representations across different societal groups \cite{rekabsaz2021societal,bigdeli2022light,ekstrand2022fairness}. Societal bias (unfairness) may reinforce negative stereotypes and perpetuate inequities in the representation of groups \cite{kay2015unequal,wu2022joint}. A specific type of societal bias is the biased representation of genders in ranked lists of documents.
Prior work on binary gender bias in document ranking associates each group ($\mathit{female}$, $\mathit{male}$) with a predefined set of gender-representative terms \cite{bigdeli2021orthogonality,rekabsaz2021societal,rekabsaz2020neural}, and measures the inequality of representation between the genders in the result list using these groups of terms. While there have been efforts in optimizing rankers for mitigating gender bias \cite{zerveas2022mitigating,rekabsaz2021societal,seyedsalehi2022bias}, there is limited research addressing the metrics that are used for the evaluation of this bias. The commonly used metrics for gender bias evaluation are \emph{average rank bias} (which we refer to as ARB)~\cite{rekabsaz2020neural} and \emph{normalized fairness in the ranked results} (NFaiRR)~\cite{rekabsaz2021societal}. These metrics have been found to result in inconsistent fairness evaluation results~\cite{klasnja2022characteristics}.

There are certain characteristics of ARB and NFaiRR that limit their utility for bias evaluation of ranked result lists: ARB provides a signed and unbounded
value for each query \cite{rekabsaz2020neural}, and therefore the bias (unfairness) values are not properly comparable across queries. 
NFaiRR evaluates a ranked list by aggregating over the unbiasedness score of each ranked document.
This approach may result in problematic evaluation results. 
Consider Figure~\ref{fig:sample-nfair}, which shows two rankings for a single query where the unbiasedness score of all documents is zero (as each document is completely biased to one group). 
The fairness of these two rankings in terms of NFaiRR is zero (i.e., both have minimum fairness), while it is intuitively clear that the ranking on the left is fairer as it provides a more balanced representation of the two groups. 
\begin{table} [t]
      \centering
      \renewcommand{\arraystretch}{0.1}
        \scalebox{0.64}{%
        \begin{tabular}{p{0.5cm}p{8.8cm}p{0.5cm}p{8.8cm}}
           \toprule 
           \multicolumn{4}{c}{Query: Who is the best football player}  
            \\ 
           \midrule
           \circled{1}  & \cellcolor{pink} ... currently he plays for Ligue 1 club Paris Saint-Germain ... &  \circled{1} & \cellcolor{pink} ... currently he plays for Ligue 1 club Paris Saint-Germain ... 
            \\  
            \midrule[0mm]
            \circled{2} & \cellcolor{skyblue} ... she previously played for Espanyol and Levante ... & \circled{2} & \cellcolor{pink} ... He is Real Madrid's all-time top goalscorer, scoring 451 ...
            \\ 
            \midrule[0mm]
            \circled{3} &  \cellcolor{skyblue}... She became the first player in the history of the league ... & \circled{3} & \cellcolor{pink} ...  he was named the Ligue 1 Player of the Year, selected to ...
            \\ 
            \midrule[0mm]
            \circled{4} & \cellcolor{pink}... he returned to Manchester United in 2021 after 12 years ... & \circled{4} & \cellcolor{pink}... he returned to Manchester United in 2021 after 12 years ... 
            \\  
            \bottomrule
        \end{tabular}}
        \captionof{figure}{Two ranked lists of retrieved results for ``who is the best football player''. Documents in blue contain only female-representative terms and documents in red contain only male-representative terms. In terms of NFaiRR, fairness of both ranked result lists is zero (minimum fairness).}
        \label{fig:sample-nfair}
\end{table}
There are metrics, however, that are not prone to the kind of problematic cases shown in Figure~\ref{fig:sample-nfair}, but are not directly applicable to fairness evaluation based on term-based group representation off-the-shelf. 
In particular, \emph{attention-weighted rank fairness} (AWRF) \cite{sapiezynski2019quantifying,raj2020comparing,trec-fair-ranking-2021} works based on soft attribution of items (here, documents) to multiple groups.
AWRF is a generic metric; for a specific instantiation it requires definitions of:
\begin{enumerate*}[label=(\roman*)] 
\item the association of items of a ranked list with respect to each group, 
\item a weighting schema, which determines the weights for different rank positions, 
\item the target distribution of groups, and 
\item a distance function to measure the difference between the target distribution of groups with their distribution in the ranked list.
\end{enumerate*} 
We propose a new metric \emph{TExFAIR} (term exposure-based fairness) based on the AWRF framework for measuring fairness of the representation of different groups in a ranked list. 
TExFAIR extends AWRF with two adaptations: 
\begin{enumerate*}[label=(\roman*)] 
\item an explicit definition of the association of documents to groups based on probabilistic \mbox{term}-level associations, and 
\item a ranked-biased discounting factor (RBDF) for counting non-representative documents towards the measurement of the fairness of a ranked list.
\end{enumerate*}

Specifically, we define the concept of \emph{term exposure} as the amount of attention each \emph{term} receives in a ranked list, given a query and a retrieval system. 
Using term exposure of group-representative terms, we estimate the extent to which each group is represented in a ranked result list. We then leverage the discrepancy in the representation of different groups to measure the degree of fairness in the ranked result list. Moreover, we show that the estimation of fairness may be highly impacted by whether the non-representative documents (documents that do not belong to any of the groups) are taken into account or not. To count these documents towards the estimation of fairness, we propose a rank-biased discounting factor (RBDF) in our evaluation metric. 
Finally, we employ counterfactual data substitution (CDS) \cite{maudslay2019s} to measure the gender sensitivity of a ranking model in terms of the discrepancy between its original rankings and the ones it provides if it performs retrieval in a counterfactual setting, where the gender of each gendered term in the documents of the collection is reversed, e.g., ``he'' $\to$ ``she,'' ``son'' $\to$ ``daughter.'' 

In summary, our main contributions are as follows:
\begin{itemize}[leftmargin=*]
\item We define an extension of the AWRF evaluation framework with the metric \emph{TExFAIR}, which explicitly defines the association of each document to the groups based on a probabilistic term-level association.
\item We show that non-representative documents, i.e., documents without any representative terms, may have a high impact in the evaluation of fairness with group-repre\-sentative terms and to address this issue we define a rank-biased discounting factor (RBDF) in our proposed metric.
\item We evaluate a set of ranking models in terms of gender bias and show that
the correlation between TExFAIR and NFaiRR is not strong, indicating that TExFAIR measures a different dimension of fairness than NFaiRR.
\end{itemize}

\section{Background} \label{sec:background}
\vspace*{-2mm}
\textbf{Fairness in rankings.}
Fairness is a subjective and context-specific constraint and there is no unique definition when it comes to defining fairness for rankings \cite{mcdonald2022search,singh2018fairness,zehlike2022fairness,abolghasemi2023retbias,heuss2022fairness}. The focus of this paper is on measuring fairness in the representation of groups in rankings \cite{rekabsaz2021societal,raj2022measuring,gao2020toward,morik2020controlling,zehlike2020reducing}, and, specifically, the setting in which each group can be represented by a predefined set of group-representative terms. 
We particularly investigate gender bias in document ranking and follow  prior work \cite{rekabsaz2020neural, rekabsaz2021societal, zerveas2022mitigating, bigdeli2023biasing,heuss2023} on gender bias in the binary setting of two groups: female and male.
In this setup, each gender is defined by a set of gender-representative terms (words), which we adopt from prior work \cite{rekabsaz2021societal}.

Previous studies on evaluating gender bias \cite{rekabsaz2020neural,rekabsaz2021societal,bigdeli2022light,zerveas2022mitigating} mostly use the ARB \cite{rekabsaz2020neural} and NFaiRR \cite{rekabsaz2021societal} metrics.
Since the ARB metric has undesirable properties (e.g., being unbounded), for the purposes of this paper we will focus on comparing our newly proposed metric to NFaiRR as the most used and most recent of the two metrics~\cite{rekabsaz2021societal,zerveas2022mitigating}. Additionally, there is a body of prior work addressing the evaluation of fairness based on different aspects \cite{yang2017measuring, zehlike2017fa,biega2018equityamortized,singh2018fairness,diaz2020evaluating,ghosh2021fair}. The metrics used in these works vary in different dimensions including 
\begin{enumerate*}[label=(\roman*)]
\item the goal of fairness, i.e., what does it mean to be fair, 
\item whether the metric considers relevance score as part of the fairness evaluation, 
\item binary or non-binary group association of each document, 
\item the weighting decay factor for different positions, and 
\item evaluation of fairness in an individual ranked list or multiple rankings~\cite{raj2020comparing,raj2022measuring}. 
\end{enumerate*}
In light of the sensitivity of gender fairness, which poses a constraint where each ranked list is supposed to represent different gender groups in a ranked list equally \cite{rekabsaz2021societal,zerveas2022mitigating,bigdeli2022light}, we adopt attention-weighted rank fairness (AWRF) \cite{sapiezynski2019quantifying} as a framework for the evaluation of group fairness in an \emph{individual ranked list} with soft attribution of documents to multiple groups.
\header{Normalized fairness of retrieval results (NFaiRR)} 
In the following, $q$ is a query, $\operatorname{tf}(t,d)$ stands for the frequency of term $t$ in document $d$, $G$ is the set of $N$ groups where $G_i$ is the $i$-{th} group with $i \in \{1, \ldots, N\}$, $V_{G_i}$ is the set of group-representative terms for group $G_i$, $d_q^r$ is the retrieved document at rank $r$ for query $q$, and $k$ is the ranking cut-off. $M^{G_i}(d)$ represents the magnitude of group $G_i$, which is equal to the frequency of $G_i$'s representative terms in document $d$,
i.e., $\mathit{M^{G_i}(d) = \sum_{t \in V_{G_i}}^{}{\operatorname{tf}(t,d)}}$. $\tau$ sets a threshold for considering a document as neutral based on $\mathit{M^{G_i}(d)}$ of all groups in $G$. Finally, $J_{G_i}$ is the expected proportion of group $G_i$ in a balanced representation of groups in a document, e.g., $J_{G_i}=\frac{1}{2}$ in equal representation for $G_i\in \{\mathit{female},\mathit{male}\}$ \cite{rekabsaz2021societal,bigdeli2022light,zerveas2022mitigating}.

Depending on $M^{G_i}(d)$ for all $G_i \in G$, document $d$ is assigned with a neutrality (unbiasedness) score $\omega(d)$:
\begin{align}
  \omega(d)&=
    \begin{cases}
      1, & \mbox{}\hspace*{-2mm}\text{if } \sum\limits_{G_i \in G}^{}{ M^{G_i}(d)\leq\tau}
      \hspace*{-2mm}\mbox{}\\
      1-\sum\limits_{G_i \in G}^{}{\abs*{\frac{M^{G_i}(d)}{\sum\limits_{G_x \in G}^{}{M^{G_x}(d)}}\!-\!J_{G_i}}}, & \mbox{}\hspace*{-2mm}\text{otherwise.}
    \end{cases}  
    \label{eq:nfairr1}
\end{align}
To estimate the fairness of the top-$k$ documents retrieved for query $q$, first, the neutrality score of each ranked document $d_q^r$ is discounted with its corresponding position bias, i.e., $(\log (r+1))^{-1}$, and then, an aggregation over top-$k$ documents is applied (Eq. \ref{eq:nfairr2}). The resulting score is referred to as the \emph{fairness of retrieval results} (FaiRR) for query $q$:
\begin{equation}
    \operatorname{FaiRR}(q,k) = \sum_{r=1}^{k}{\frac{\omega(d_q^r)}{\log(r+1)}}.
     \label{eq:nfairr2}
\end{equation}
As FaiRR scores of different queries may end up in different value ranges (and consequently are not comparable across queries), a background set of documents $S$ is employed to normalize the fairness scores with the \emph{ideal FaiRR} (IFaiRR) of $S$ for query $q$ \cite{rekabsaz2021societal}. $\operatorname{IFaiRR}(q,S)$ is the best possible fairness result that can be achieved from
reordering the documents in the background set $S$ \cite{rekabsaz2021societal}. The NFaiRR score for a query is formulated as follows:
\begin{equation}
    \label{eq:nfairr3}
    \operatorname{NFaiRR}(q,k,S) = \frac{\operatorname{FaiRR}(q,k)} {\operatorname{IFaiRR}(q,S)}.
\end{equation}
%

\header{Attention-weighted rank fairness (AWRF)} 
\label{subsec:AWRF}
Initially proposed by Sapiezynski et al. \cite{sapiezynski2019quantifying}, AWRF measures the unfairness of a ranked list based on the difference between the exposure of groups and their target exposure. To this end, it first computes a vector $E_{L_q}$ of the accumulated exposure that a list of $k$ documents $L$ retrieved for query $q$ gives to each group:
\begin{equation}
    E_{L_q} = \sum_{r=1}^{k}{v_r a_{d_q^r}}.
\end{equation}
Here, $v_r$ represents the attention weight, i.e., position bias corresponding to the rank $r$, e.g., $(\log (r+1))^{-1}$ \cite{sapiezynski2019quantifying,trec-fair-ranking-2021}, and $a_{d_q^r} \in [0,1]^{|G|} $ stands for the alignment vector of document $d_q^r$ with respect to different groups in the set of all groups $G$. Each entity in the alignment vector $a_{d_q^r}$ determines the association of $d_q^r$ to one group, i.e., $a_{d_q^r}^{G_i}$.
To convert $E_{L_q}$ to a distribution, a normalization is applied:
\begin{equation}
    nE_{L_q} = \frac{E_{L_q}}{\Vert E_{L_q}\Vert_1}. 
    \label{eq:awrf-normalization}
\end{equation}
Finally, a distance metric is employed to measure the difference between the desired target distribution $\hat{E}$ and the $nE_{L_q}$, the distribution of groups in the ranked list retrieved for query $q$:
\begin{equation}
    \label{eq:awrf-distance}
    \operatorname{AWRF}(L_q)=\Delta(nE_{L_q},\hat{E}).
\end{equation}

\section{Methodology}
\label{sec:methodology}

As explained in Section~\ref{sec:introduction} and~\ref{sec:background}, NFaiRR measures fairness based on document-level unbiasedness scores.
However, in measuring the fairness of a ranked list, individual documents might be biased while the ranked list as a whole balances the groups' representations.
Hence, fairness in the representation of groups in a ranked list should not be defined as an aggregation of document-level scores. 

We, therefore, propose to measure group representation for a top-$k$ ranking using term exposure in the ranked list as a whole. We adopt the weighting approach of $\operatorname{AWRF}$, and explicitly define the association of documents on a term-level.
Additionally, as we show in Section \ref{sec:results}, the effect of documents without any group-representative terms, i.e., non-representative documents, could result in under-estimating the fairness of ranked lists. To address this issue, we introduce a rank-biased discounting factor in our metric.
Other measures for group fairness exist, and some of these measures also make use of exposure \cite{diaz2020evaluating,singh2018fairness}.\footnote{Referring to the amount of attention an item (document) receives from users in the ranking.}  However, these measures are not at the term-level, but at the document-level. In contrast, we perform a finer measurement and quantify the amount of attention a \textit{term} (instead of document) receives.

\header{Term exposure}
In order to quantify the amount of attention a specific term $t$ receives given a ranked list of $k$ documents retrieved for a query $q$, we formally define \emph{term exposure} of term $t$ in the list of $k$ documents $L_q$ as follows:
\begin{equation}
    \label{eq:term-exposure1}
    \mbox{}\hspace*{-2mm}
    \operatorname{TE}@k(t,q,L_q)= \sum_{r=1 }^{k}{p_o(t\mid d_q^r)\cdot p_o(d_q^r)}.
\end{equation}
Here, $d^r_q$ is a document ranked at rank $r$ in the ranked result retrieved for query $q$. 
$p_o(t\mid d_q^r)$ is the probability of observing term $t$ in document $d_q^r$, and $p_o(d_q^r)$ is the probability of document $d$ at rank $r$ being observed by user. We can perceive $p_o(t\mid d_q^r)$ as the probability of term $t$ occurring in document $d_q^r$. Therefore, using maximum likelihood estimation, we estimate $p_o(t\mid d_q^r)$ with the frequency of term $t$ in document $d_q^r$ divided by the total number of terms in $d_q^r$, i.e., $\operatorname{tf}(t,d_q^r)\cdot \vert d_q^r \vert^{-1}$. Additionally, following~\cite{singh2018fairness,mcdonald2022search}, we assume that the observation probability $p_o(d_q^r)$ only depends on the rank position of the document, and therefore can be estimated using the position bias at rank $r$. Following~\cite{singh2018fairness,rekabsaz2021societal}, we define the position bias as $(\log (r+1))^{-1}$. Accordingly, Eq.~\ref{eq:term-exposure1} can be reformulated as follows:
\begin{equation}
    \label{eq:term-exposure2}
    \operatorname{TE}@k(t,q)= \sum_{r=1 }^{k}{\frac{\frac{\operatorname{tf}(t,d_q^r)}{\vert d_q^r\vert}}{\log (r+1)}}.
\end{equation}

\header{Group representation}
We leverage the term exposure (Eq. \ref{eq:term-exposure2}) to estimate the representation of each group using the exposure of its representative terms as follows:
\begin{equation}
    \label{eq:group-representation}
        p(G_i\mid q,k) = \frac{\sum_{t \in V_{G_i}}^{}{\operatorname{TE}@k(t,q)}}{\sum_{G_x \in G}^{}{\sum_{t \in V_{G_x}}^{}{\operatorname{TE}@k(t,q)}}}.
\end{equation}
Here, $G_i$ represents the group $i$ in the set of $N$ groups indicated with $G$ (e.g., $G=\{\mathit{female}, \mathit{male}\}$), and $V_{G_i}$ stands for the set of terms representing group $G_i$. The component ${\sum_{G_x \in G}^{}{\sum_{t \in V_{G_x}}^{}{\operatorname{TE}@k(t,q)}}} $ can be interpreted as the total amount of attention that users spend on the representative terms in the ranking for query $q$.
This formulation of the group representation corresponds to the normalization step in AWRF (Eq.~\ref{eq:awrf-normalization}).

\header{Term exposure-based divergence}
\label{subsec:texfair}
To evaluate the fairness based on the representation of different groups, we define a fairness criterion built upon our term-level perspective in the representation of groups: in a fairer ranking -- one that is less biased -- each group of terms receives an amount of attention proportional to their corresponding desired target representation. Put differently, a divergence from the target representations of groups can be used as a means to measure the bias in the ranking. This divergence corresponds to the distance function in Eq. \ref{eq:awrf-distance}. Let $\hat{p}_{Gi}$ be the target group representation for each group $G_i$ (e.g., $\hat{p}_{G_i}=\frac{1}{2}$ for $G_i\in\{\mathit{female}, \mathit{male}\}$ for equal representation of male and female), then we can compute the bias in the ranked results retrieved for the query $q$ as the absolute divergence between the groups' representation and their corresponding target representation. We refer to this bias as the \emph{term exposure-based divergence} (TED) for query $q$:
\begin{equation}
    \label{eq:ted1}
    \operatorname{TED}(q, k)= {\sum_{G_i \in G}^{}{| p(G_i\mid q,k)-\hat{p}_{G_i} |}}.
\end{equation}

\header{Rank-biased discounting factor (RBDF)}
With the current formulation of group representation in Eq.~\ref{eq:group-representation}, non-representative documents, i.e., the documents that do not include any group-representative terms, will not contribute to the estimation of bias in TED (Eq. \ref{eq:ted1}). To address this issue, we discount the bias in Eq. \ref{eq:ted1} with the proportionality of those documents that count towards the bias estimation, i.e., documents which include at least one group-representative term. To take into account each of these documents with respect to their position in the ranked list, we leverage their corresponding position bias, i.e., $(\log(1+r))^{-1}$ for a document at rank $r$, to compute the proportionality. The resulting proportionality factor which we refer to as \emph{rank-biased discounting factor} (RBDF) is estimated as follows:
\begin{equation}
    \label{eq:RDBF}
    \operatorname{RBDF}(q,k) = \frac{\sum_{r=1}^{k}{\frac{\mathds{1}[d_q^r \in S_R]}{\log (1+r) }}}{\sum_{r=1}^{k}{\frac{1}{\log (1+r)}}}.
\end{equation}
Here, $S_R$ stands for the set of representative documents in top-$k$ ranked list of query $q$, i.e., documents that include at least one group-representative term. Besides, $\mathds{1}[d_q^r \in S_R]$ is equal to 1 if $d_q^r \in S_R$, otherwise, 0. Accordingly, we incorporate $\operatorname{RDBF}(q,k)$ into Eq.~\ref{eq:ted1} and reformulate it as:
\begin{equation}
    \operatorname{TED}(q, k) = {\sum_{G_i \in G}^{}{| p(G_i\mid q, k)-\hat{p}_{G_i} |}} \cdot \frac{\sum_{r=1}^{k}{\frac{\mathds{1}[d_q^r \in S_R]}{\log (1+r) }}}{\sum_{r=1}^{k}{\frac{1}{\log (1+r)}}}.
    \label{eq:ted-w-prp}
\end{equation}
Alternatively, as $\operatorname{TED}(q, k)$ is bounded, we can leverage the maximum value of $\operatorname{TED}$ to quantify the fairness of the rank list of query $q$. We refer to this quantity as \emph{term exposure-based fairness} (TExFAIR) of query $q$:
\begin{equation}
    \operatorname{TExFAIR}(q, k) =  \max(\operatorname{TED})-\operatorname{TED}(q, k).
\end{equation}
In the following, we use $\operatorname{TExFAIR}$ to refer to $\operatorname{TExFAIR}$ with proportionality (RBDF), unless otherwise stated. With $\hat{p}_{G_i}=\frac{1}{2}$ for
$G_i\in \{\mathit{female},\mathit{male}\}$, $\operatorname{TED}$ (Eq. \ref{eq:ted1} and \ref{eq:ted-w-prp}) falls into the range of [0,1], therefore $\operatorname{TExFAIR}(q,k) = 1-\operatorname{TED}(q, k)$.

\section{Experimental Setup}
\header{Query sets and collection}
We use the MS MARCO Passage Ranking collection \cite{bajaj2016ms}, and evaluate the fairness on two sets of queries from prior work \cite{zerveas2022mitigating,rekabsaz2020neural,bigdeli2022light}: 
\begin{enumerate*}[label=(\roman*)]
\item QS1 which consists of 1756 non-gendered queries \cite{rekabsaz2020neural}, and 
\item QS2 which includes 215 bias-sensitive queries
\cite{rekabsaz2021societal} (see \cite{rekabsaz2020neural} and \cite{rekabsaz2021societal} respectively for examples).
\end{enumerate*}
\par
\header{Ranking models}
Following the most relevant related work \cite{zerveas2022mitigating,rekabsaz2021societal}, we evaluate a set of ranking models which work based on pre-trained language models (PLMs).
Ranking with PLMs can be classified into three main categories: sparse retrieval, dense retrieval, and re-rankers. In our experiments we compare the following models: 
\begin{enumerate*}[label=(\roman*)]
\item two \emph{sparse retrieval} models:
 uniCOIL\cite{lin2021unicoil}
and DeepImpact \cite{mallia2021learning}; 
\item five \emph{dense retrieval} models: ANCE \cite{xiong2020approximateANCE}, TCT-ColBERTv1 \cite{lin2020tctcolbertv1}, SBERT \cite{reimers-2019sbert}, distilBERT-KD \cite{hofstatter2020distilkd}, and distilBERT-TASB \cite{hofstatter2021distilTas}; 
\item three commonly used cross-encoder \emph{re-rankers}: BERT \cite{nogueira2019passage}, MiniLM$_\mathit{KD}$ \cite{wang2020minilm} and TinyBERT$_{KD}$ \cite{jiao2020tinybert}. Additionally, we evaluate BM25 \cite{robertson1994some} as a widely-used traditional lexical ranker \cite{abolghasemi2022interpolation,lin2021pretrained}.
\end{enumerate*}
%
For sparse and dense retrieval models we employ the pre-built indexes, and their corresponding query encoders provided by the Pyserini toolkit \cite{lin2021pyserini}.
For re-rankers, we use the pre-trained cross-encoders provided by the sentence-transformers library \cite{reimers-2019sbert}.\footnote{\url{https://www.sbert.net/docs/pretrained-models/ce-msmarco.html}}
For ease of fairness evaluation in future work, we make our code publicly available at https://github.com/aminvenv/texfair.

\header{Evaluation details}
We use the official code available for NFaiRR.\footnote{\url{https://github.com/CPJKU/FairnessRetrievalResults}} Following suggestions in prior work~\cite{rekabsaz2021societal}, we utilize the whole collection as the background set $S$ (Eq. \ref{eq:nfairr3}) to be able to do the comparison across rankers and re-rankers (which re-rank top-1000 passages from BM25). Since previous instantiations of AWRF cannot be used for the evaluation of term-based fairness of group representations out-of-the-box, we compare TExFAIR to NFaiRR.

\section{Results}
\label{sec:results}
\begin{table*}[t]
        \centering
        \caption{Effectiveness and fairness results at ranking cut-off $=10$.
        $r$ denotes the correlation between TExFAIR and NFaiRR. Higher values of TExFAIR and NFaiRR correspond to higher fairness. $\dagger$ denotes statistical significance for correlations with  $(p<0.05)$. $\ddagger$ indicates statistically significant improvement over BM25 according to a paired t-test ($p<0.05$). Bonferroni correction is used for multiple testing.}
        \renewcommand{\arraystretch}{1.27}
        \setlength{\tabcolsep}{2.1pt}
        \label{tab:bias-effectiveness}
        \scalebox{0.775}{
        \begin{tabular}{l ccccccc ccccccc}
            \toprule
            \multirow{2}{*}{\textbf{Method}} 
            &
            \multicolumn{5}{c}{QS1} &
            \multicolumn{5}{c}{QS2}
            \\
            \cmidrule(r){2-6}
            \cmidrule{7-11}
            &
            {MRR} & {nDCG} &
            {NFAIRR} &
            {TExFAIR} &
            {$r$} &
            {MRR}& {nDCG}&
            NFAIRR{} &
            {TExFAIR} &
            {$r$}
            \\
            \midrule
            \textbf{Sparse retrieval} &
            
            &  &  & & & & &
            \\ 
            \;  BM25 &
            0.1544 & 0.1958 &
            0.7227  & 
            0.7475 &
            0.4823\rlap{$^{\dagger}$} &
            0.0937 & 0.1252 &
            0.8069 &
            0.8454 &
            0.5237\rlap{$^{\dagger}$}
              \\           
            \; UniCOIL &          
            0.3276\rlap{$^{\ddagger}$} & 0.3892\rlap{$^{\ddagger}$} &
            0.7819\rlap{$^{\ddagger}$}  & 
            0.7629\rlap{$^{\ddagger}$} &
            0.5166\rlap{$^{\dagger}$} &
            0.2288\rlap{$^{\ddagger}$} & 0.2726\rlap{$^{\ddagger}$} &
            0.8930\rlap{$^{\ddagger}$} &
            0.8851\rlap{$^{\ddagger}$} & 
            0.4049\rlap{$^{\dagger}$}   \\  
              \; DeepImpact &
            0.2690\rlap{$^{\ddagger}$} &  0.3266\rlap{$^{\ddagger}$} &
            0.7721\rlap{$^{\ddagger}$}  & 
            0.7633\rlap{$^{\ddagger}$} &
            0.5487\rlap{$^{\dagger}$} &
            0.1788\rlap{$^{\ddagger}$} & 0.2200\rlap{$^{\ddagger}$} &
            0.8825\rlap{$^{\ddagger}$} &
            0.8851\rlap{$^{\ddagger}$} & 
            0.4971\rlap{$^{\dagger}$}
            \\
            \midrule

            \textbf{Dense retrieval} &
             &  &
             &  &
             & &  &
            \\
            \;  ANCE &
            0.3056\rlap{$^{\ddagger}$} &  0.3640\rlap{$^{\ddagger}$} &
            \textbf{0.7989}\rlap{$^{\ddagger}$} & 
            \textbf{0.7725}\rlap{$^{\ddagger}$} &
             0.5181\rlap{$^{\dagger}$} &
            0.2284\rlap{$^{\ddagger}$} &  0.2763\rlap{$^{\ddagger}$} &
            0.9093\rlap{$^{\ddagger}$} &
            \textbf{0.9060}\rlap{$^{\ddagger}$} &
            0.4161\rlap{$^{\dagger}$}
            \\ 
            \; DistillBERT$_\mathit{KD}$ &
            
            0.2906\rlap{$^{\ddagger}$} & 0.3488\rlap{$^{\ddagger}$} &
            0.7913\rlap{$^{\ddagger}$}  & 
            0.7683\rlap{$^{\ddagger}$} &
            0.5525\rlap{$^{\dagger}$} &
            0.2306\rlap{$^{\ddagger}$} & 0.2653\rlap{$^{\ddagger}$} &
            \textbf{0.9149}\rlap{$^{\ddagger}$} &
            0.9044\rlap{$^{\ddagger}$} &
            0.4257\rlap{$^{\dagger}$}
            \\ 
            \; 
            DistillBERT$_\mathit{TASB}$ &
            0.3209\rlap{$^{\ddagger}$} &  0.3851\rlap{$^{\ddagger}$} &
            0.7898\rlap{$^{\ddagger}$}  & 
            0.7613\rlap{$^{\ddagger}$} &
            0.5091\rlap{$^{\dagger}$} &
            0.2250\rlap{$^{\ddagger}$} & 0.2725\rlap{$^{\ddagger}$} &
            0.9088\rlap{$^{\ddagger}$} &
            0.8960\rlap{$^{\ddagger}$} &
            0.4073\rlap{$^{\dagger}$}
            \\
            \; TCT-ColBERTv1 &
            0.3138\rlap{$^{\ddagger}$} &  0.3712\rlap{$^{\ddagger}$} &
            0.7962\rlap{$^{\ddagger}$}  & 
            0.7688\rlap{$^{\ddagger}$} &
             0.5253\rlap{$^{\dagger}$} &
            0.2300\rlap{$^{\ddagger}$} & 0.2732\rlap{$^{\ddagger}$} &
            0.9116\rlap{$^{\ddagger}$} &
            0.9056\rlap{$^{\ddagger}$} &
            0.4249\rlap{$^{\dagger}$}
            \\ 
            \; SBERT &
            0.3104\rlap{$^{\ddagger}$} &  0.3693\rlap{$^{\ddagger}$} &
            0.7880\rlap{$^{\ddagger}$}  & 
            0.7637\rlap{$^{\ddagger}$} &
            0.5217\rlap{$^{\dagger}$} &
            0.2197\rlap{$^{\ddagger}$} & 0.2638\rlap{$^{\ddagger}$} &
            0.8943\rlap{$^{\ddagger}$} &
            0.8999\rlap{$^{\ddagger}$} &
             0.3438\rlap{$^{\dagger}$}
            \\   
            \midrule

            \textbf{Re-rankers} &

             &  &
             &  &
             & &
             & &  &
            \\
            \; BERT &
            0.3415\rlap{$^{\ddagger}$} &  0.4022\rlap{$^{\ddagger}$} &
           0.7790\rlap{$^{\ddagger}$} &
            0.7584\rlap{$^{\ddagger}$} &
            0.5135\rlap{$^{\dagger}$} &
            0.2548\rlap{$^{\ddagger}$} & 0.2950\rlap{$^{\ddagger}$} &
            0.8896\rlap{$^{\ddagger}$} &
            0.8807\rlap{$^{\ddagger}$} &
            0.4323\rlap{$^{\dagger}$}
            \\
            \; MiniLM$_\mathit{KD}$  &
            \textbf{0.3832}\rlap{$^{\ddagger}$} &  \textbf{0.4402}\rlap{$^{\ddagger}$} &
           0.7702\rlap{$^{\ddagger}$}  &
            0.7516 &
            0.5257\rlap{$^{\dagger}$} &
            \textbf{0.2872}\rlap{$^{\ddagger}$} &  \textbf{0.3323}\rlap{$^{\ddagger}$} &
            0.8863\rlap{$^{\ddagger}$} &
            0.8865\rlap{$^{\ddagger}$} &
            0.3880\rlap{$^{\dagger}$}
            \\
            \; TinyBERT$_\mathit{KD}$ &
            0.3482\rlap{$^{\ddagger}$} &  0.4093\rlap{$^{\ddagger}$} &
            0.7799\rlap{$^{\ddagger}$} & 
            0.7645\rlap{$^{\ddagger}$} &
            0.5437\rlap{$^{\dagger}$} &
            0.2485\rlap{$^{\ddagger}$} & 0.3011\rlap{$^{\ddagger}$} &
            
            0.8848\rlap{$^{\ddagger}$} &
            0.8952\rlap{$^{\ddagger}$} &
            0.4039\rlap{$^{\dagger}$}
            \\
            \bottomrule
    \end{tabular}
    }
    \end{table*} 
Table \ref{tab:bias-effectiveness} shows the evaluation of the rankers in terms of effectiveness (MRR and nDCG) and fairness (NFaiRR and TExFAIR). The table shows that almost all PLM-based rankers are significantly fairer than BM25 on both query sets at ranking cut-off 10.
In the remainder of this section we address three questions:
\begin{enumerate}[label=(\roman*)]
\item What is the correlation between the proposed TExFAIR metric and the commonly used NFaiRR metric?
\item What is the sensitivity of the metrics to the ranking cut-off? 
\item What is the relationship between the bias in ranked result lists of rankers, and how sensitive they are towards the concept of gender?

\end{enumerate}

\header{(i) Correlation between metrics}
To investigate the correlation between the TExFAIR and NFaiRR metrics, we employ Pearson's correlation coefficient on the query level. As the values in Table~\ref{tab:bias-effectiveness} indicate, the two metrics are significantly correlated, but the relationship is not strong ($0.34<r<0.55$). This is likely due to the fact that NFaiRR and TExFAIR are structurally different: NFaiRR is document-centric: it estimates the fairness in the representation of groups on a document-level and then aggregates the fairness values over top-$k$ documents. TExFAIR, on the other hand, is ranking-centric: each group's representation is measured based on the whole ranking, instead of individual documents. As a result, in a ranked list of $k$ documents, the occurrences of the terms from one group at rank $i$, with $i\in\{1,\ldots,k\}$, can balance and make up for the occurrences of the other group's terms at rank $j$, with $j\in\{1,\ldots,k\}$. This is in contrast to NFaiRR in which the occurrences of the terms from one group at rank $i$, with $i\in\{1,\ldots,k\}$, can only balance and make up for the occurrences of other group's terms at rank $i$. 
Thus, TExFAIR measures a different dimension of fairness than NFaiRR. 

\begin{figure*}[t]
    \centering
    \includegraphics[width=\columnwidth]{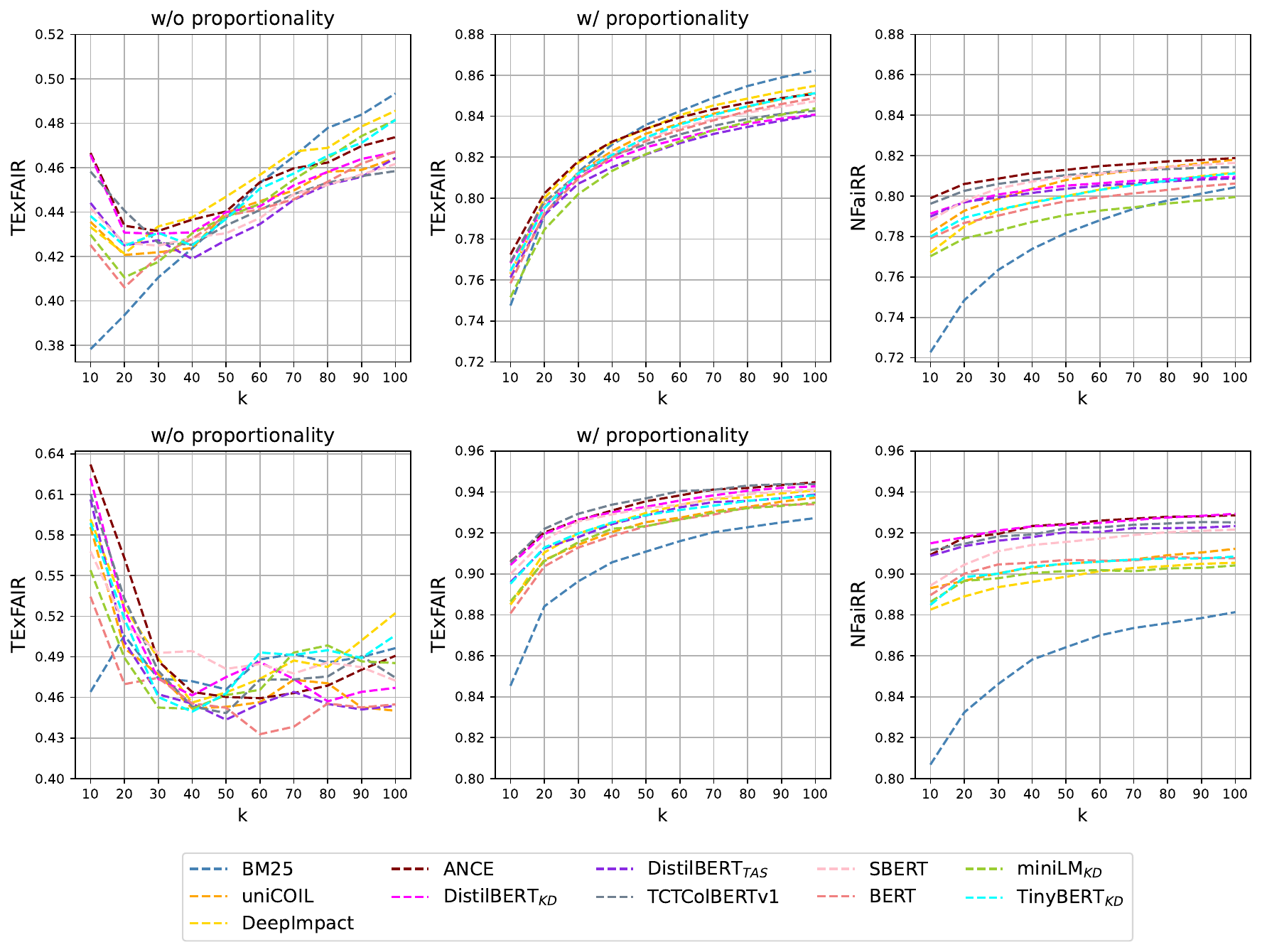}
    \caption{Fairness results on QS1 (first row) and QS2 (second row) at different ranking cut-off values ($k$).}
    \label{fig:tef-nfairr-result}
\end{figure*}

\header{(ii) Sensitivity to ranking cut-off $k$}
Figure \ref{fig:tef-nfairr-result} depicts the fairness results at various cut-offs using TExFAIR with and without proportionality (RBDF) as well as the results using NFaiRR. The results using TExFAIR without proportionality show a high sensitivity to the ranking cut-off $k$ in comparison to the other two metrics.
The reason is that without proportionality factor RDBF, the unbiased documents with zero group-representative term, i.e., non-representative documents, do not count towards the fairness evaluation. 
As a result, regardless of the number of  this kind of unbiased documents, documents that include group-representative terms potentially can highly affect the fairness of the ranked list. On the contrary, NFaiRR and TExFAIR with proportionality factor are less sensitive to the ranking cut-off: the effect of unbiased documents with zero group-representative term is addressed in NFaiRR with a maximum neutrality for these documents (Eq. \ref{eq:nfairr1}), and in TExFAIR with proportionality factor RBDF by discounting the bias using the proportion of documents that include group-representative terms (Eq. \ref{eq:ted-w-prp}). 

\header{(iii) Counterfactual evaluation of gender fairness}
TExFAIR and NFaiRR both measure the fairness of ranked lists produced by ranking models. Next, we perform an analytical evaluation to measure the extent to which a ranking model acts indifferently (unbiasedly) towards the genders, regardless of the fairness of the ranked list it provides. Our evaluation is related to counterfactual fairness measurements which require that the same
outcome should be achieved in the real world as in the term-based counterfactual world \cite{pearl2009causal,wu2019counterfactual}. Here, the results of the real world correspond to the ranked lists that are returned using the original documents, and results of the counterfactual world correspond to the ranked lists that are returned using counterfactual documents. 
\par
In order to construct counterfactual documents, we employ counterfactual data substitution (CDS) \cite{lu2020gender,maudslay2019s}, in which we replace terms in the collection with their counterpart in the opposite gender-representative terms, e.g., ``he'' $\to$ ``she,'' ``son'' $\to$ ``daughter,'' etc. For names, e.g., Elizabeth or John, we substitute them with a name from the opposite gender name in the gender-representative terms \cite{maudslay2019s}. Additionally, we utilize POS information to avoid ungrammatically assigning ``her'' as a personal pronoun or possessive determiner \cite{maudslay2019s}.
We then measure how the ranked result lists of a ranking model on a query set $Q$ would diverge if a ranker performs the retrieval on the counterfactual collection rather than the original collection.
\par
In order to measure the divergence, we employ rank-biased overlap (RBO) \cite{webber2010similarity} as a measure to quantify the similarities between two ranked lists. 
We refer to this quantity as \emph{counterfactually-estimated rank-biased overlap} (CRBO). RBO ranges from 0 to 1, where 0 represents disjoint and 1 represents identical ranked lists. RBO has a parameter $0<p\leq1$ which regulates the degree of top-weightedness in estimating the similarity. From another perspective, $p$ represents searcher patience or persistence and larger values of $p$ stand for more persistent searching \cite{clarke2021assessing}. Since we focus on top-10 ranked results, we follow the original work \cite{webber2010similarity} for a reasonable choice of $p$, and set it to 0.9 (see \cite{webber2010similarity} for more discussion).

Table \ref{tab:crbo} shows the CRBO results.
While there is a substantial difference in the fairness of ranked results between the BM25 and the PLM-based rankers, the CRBO results of these models are highly comparable, and even BM25, as the model which provides the most biased ranked results, is the least biased model in terms of CRBO on QS1. 
Additionally, among PLM-based rankers, the ones with higher TExFAIR or NFaiRR scores do not necessarily provide higher CRBO. This discrepancy between \{NFaiRR, TExFAIR\} and CRBO disentangles the bias of a model towards genders from the bias of the ranked results it provides. However, it should be noted that we indeed cannot come to a conclusion as to whether the bias that exists in the PLM-based rankers (the one that is reflected by CRBO) does not contribute to their superior fairness of ranked results (the one that is reflected by \{NFaiRR, TExFAIR\}).
We leave further investigation of the quantification of inherent bias of PLM-based rankers and its relation with the bias of their ranked results for future work.

 \begin{table}[t]
    \centering
    \caption{Counterfactually-estimated RBO results. For ease of comparison, TExFAIR and NFaiRR results are included from Table \ref{tab:bias-effectiveness}.
    }
    \renewcommand{\arraystretch}{1.1}
    \setlength{\tabcolsep}{3pt}
    \label{tab:crbo}
    \scalebox{0.9}{
        \begin{tabular}{l ccc ccc}
        \toprule
        & \multicolumn{3}{c}{QS1} & \multicolumn{3}{c}{QS2}
        \\
        \cmidrule(r){2-4}\cmidrule{5-7}
        Models & CRBO & TExFAIR & NFaiRR & CRBO & TExFAIR & NFaiRR
        \\
        \midrule
        BM25
        & \textbf{0.9733} 
        & 0.8454
        & 0.8069 
        & \textbf{0.9761} 
        & 0.7475
        & 0.7227 
        \\
        BERT 
        & 0.9506 
        & 0.8807 
        & \textbf{0.8896}
        & 0.9735 
        & 0.7629 & 0.7790  
        \\ 
        MiniLM 
        & 0.9597
        & 0.8865 
        & 0.8863
        & 0.9753
        & 0.7516 
        & 0.7702   
        \\
        TinyBERT 
        & 0.9519
        & \textbf{0.8952}
        & 0.8848
        & 0.9714
        & \textbf{0.7645} & \textbf{0.7799}\\
        \bottomrule
    \end{tabular}
    }
\end{table}

\section{Discussion}
\textbf{The role of non-representative documents.}
As explained in Section~\ref{sec:methodology}, and based on the results in Section \ref{sec:results}, discounting seems to be necessary for the evaluation of gender fairness in document ranking with group-representative terms, due to the effect of non-repre\-sentative documents. Here, one could argue that without our proposed proportionality discounting factor (Section~\ref{subsec:texfair}), it is possible to use an association value for $d_q^r$ to group $G_i$, i.e., $a_{d_q^r}^{G_i}$ in the formulation of $\operatorname{AWRF}$ (Section \ref{subsec:AWRF}) as follows:
\begin{equation}
    a_{d_q^r}^{G_i} = \frac{M^{G_i}(d_q^r)}{\sum_{G_x \in G}^{}{M^{G_x}(d_q^r)}},
    \label{eq:doc-association}
\end{equation}
and simply assign equal association for each group, e.g., $a_{d_q^r}^{G_i}=\frac{1}{2}$ for $G_i \in \{\mathit{female}$, $\mathit{male}\}$ for documents that do not contain group-representative terms, i.e., non-repre\-sentative documents. However, we argue that such formulation results in the ignorance of the frequency of group-representative terms. For instance, intuitively, a document which has only one mention of a female name as a female-representative term (therefore is completely biased towards female) and is positioned at rank $i$, cannot simply compensate and balance for a document with high frequency of male-representative names and pronouns (completely biased towards male) and is positioned at rank $i+1$. However, with the formulation of document associations in AWRF (Eq. \ref{eq:doc-association}) these two documents can roughly\footnote{As they have different position bias.} balance for each other. As such, there is a need for a fairness estimation in which the frequency of terms is better counted towards the final distribution of groups. Our proposed metric TExFAIR implicitly accounts for this effect by performing the evaluation based on term-level exposure estimation and incorporating the rank biased discounting factor RBDF.

\header{Limitations of CRBO} 
While measuring gender bias with counterfactual data substitution is widely used for natural language processing tasks \cite{maudslay2019s,rus2022closing,czarnowska2021quantifying,garg2019counterfactual}, we believe that our analysis falls short of thoroughly measuring the learned stereotypical bias. We argue that through the pre-training and fine-tuning step, specific gendered correlations could be learned in the representation space of the ranking models \cite{webster2020measuring}. For instance, the representation of the word ``nurse'' or ``babysitter'' might already be associated with female group terms. In other words, the learned association of each term to different groups (either female or male), established during pre-training or fine-tuning, is a spectrum rather than binary. As a result, these kinds of words could fall at different points of this spectrum and therefore, simply replacing a limited number of gendered-terms (which are assumed to be the two end point of this spectrum) with their corresponding counterpart in the opposite gender group, might not reflect the actual inherent bias of PLM-based rankers towards different groups of gender. Moreover, while we estimate CRBO based on the divergence of the results on the original collection and a single counterfactual collection, more stratified counterfactual setups can be studied in future work.

\header{Reflection on evaluation with term-based representations}
We acknowledge that evaluating fairness with term-based representations is limited in comparison to real-world user evaluations of fairness. However, this shortcoming exists for all natural language processing tasks where semantic evaluation from a user's perspective might not exactly match with the metrics that work based on term-based evaluation. For instance, there exists a discussion over the usage of BLEU \cite{papineni2002bleu} and ROUGE \cite{lin2004rouge} scores in the evaluation of natural language generation \cite{sulem-etal-2018-bleu,zhang2019bertscore}. Nevertheless, such an imperfect evaluation method is still of great importance due to the sensitivity of the topic of societal fairness and the impact caused by the potential consequences of unfair ranking systems. We believe that our work addresses an important aspect of evaluation in the current research in this area and plan to work on more semantic approaches of societal fairness evaluation in the future.

\section{Conclusion}
In this paper, we addressed the evaluation of societal group bias in document ranking. We pointed out an important limitation of the most commonly used group fairness metric NFaiRR, which measures fairness based on a fairness score of each ranked document. Our newly proposed metric TExFAIR integrates two extensions on top of a previously proposed generic metric AWRF: the term-based association of documents to each group, and a rank biased discounting factor that addresses the impact of non-representative documents in the ranked list. As it is structurally different, our proposed metric TExFAIR measures a different aspect of the fairness of a ranked list than NFaiRR. Hence, when fairness is taken into account in the process of model selection, e.g., with a combinatorial metric of fairness and effectiveness \cite{rekabsaz2021societal}, the difference between the two metrics TExFAIR and NFaiRR could result in a different choice of model.

In addition, we conducted a counterfactual evaluation, estimating the inherent bias of ranking models towards different groups of gender. With this analysis we show a discrepancy between the measured bias in the ranked lists (with NFaiRR or TExFAIR) on the one hand and the inherent bias in the ranking models themselves on the other hand. In this regard, for our future work, we plan to study more semantic approaches of societal fairness evaluation to obtain a better understanding of the relationship between the inherent biases of ranking models and the fairness (unbiasedness) of the ranked lists they produce. Moreover, since measuring group fairness with term-based representations of groups is limited (compared with the real-world user evaluation of fairness), we intend to work on more user-oriented methods for the measurement of societal fairness in the ranked list of documents.

\section*{Acknowledgements}
This work was supported by 
the DoSSIER project under European Union’s Horizon 2020 research and innovation program, Marie Skłodowska-Curie grant agreement No. 860721,
the Hybrid Intelligence Center, a 10-year program funded by the Dutch Ministry of Education, Culture and Science through the Netherlands Organisation for Scientific Research, \url{https://hybrid-intelligence-centre.nl}, project LESSEN with project number NWA.1389.20.183 of the research program NWA ORC 2020/21, which is (partly) financed by the Dutch Research Council (NWO),
and
the FINDHR (Fairness and Intersectional Non-Discrimination in Human Recommendation) project that received funding from the European Union’s Horizon Europe research and innovation program under grant agreement No 101070212.

All content represents the opinion of the authors, which is not necessarily shared or endorsed by their respective employers and/or sponsors.

\bibliographystyle{splncs04}
\bibliography{references}

\end{document}